\documentclass[runningheads]{llncs}
\usepackage[T1]{fontenc}
\usepackage{graphicx}
\usepackage{booktabs}
\usepackage[misc]{ifsym}

\usepackage{mwe}

\usepackage{amsmath}
\usepackage{booktabs}

\usepackage{algorithm}
\usepackage[noend]{algpseudocode}

\usepackage{subcaption}
\usepackage{multirow}
\usepackage{xcolor}
\usepackage{xspace}
\usepackage{amssymb}
\usepackage{amsfonts}
\usepackage{mathtools}
\usepackage[inline]{enumitem} %
\usepackage{pgf}
\usepackage{tikz}
\usetikzlibrary{shapes}

\newcommand{\softl}{\text{SoftLearn}\xspace}

\newcommand{\cartesianprod}{\bigtimes}

\DeclarePairedDelimiter{\grp}{(}{)}
\DeclarePairedDelimiterX{\cgrp}[2]{(}{)}{#1\delimsize\vert#2}
\DeclarePairedDelimiter{\set}{\{}{\}}
\DeclarePairedDelimiterX{\cset}[2]{\{}{\}}{#1:#2}

\newcommand{\pexp}{\mathbb{E}}
\newcommand{\dataw}{v}
\newcommand{\dataW}{V}
\newcommand{\pexpsw}[2][\weight]{\pexp_{\subsdag{#2},#1}}
\newcommand{\pexpswv}[2][f]{\pexpsw{#2}\grp{#1}}
\newcommand{\pexpm}[1]{\pexp_{#1}}
\newcommand{\pexpmv}[2][f]{\pexpm{#2}\grp{#1}}

\newcommand{\gauss}{\mathcal{N}}
\newcommand{\nodes}{\mathbf{N}}
\newcommand{\instances}{\mathcal{D}}
\newcommand{\realns}{\mathcal{R}}
\newcommand{\edges}{\mathcal{E}}
\newcommand{\edge}[2]{#1\rightarrow #2}
\newcommand{\node}[1][N]{#1}
\newcommand{\subsdag}[1]{#1\mathord{\downarrow}}
\newcommand{\distrnode}{\node[D]}
\newcommand{\sumnode}{\node[S]}
\newcommand{\prodnode}{\node[P]}
\newcommand{\childnode}{\node[C]}
\newcommand{\rootnode}{\node[R]}

\DeclareMathOperator{\ch}{ch}
\newcommand{\children}[1]{\ch{#1}}
\DeclareMathOperator{\de}{de}
\newcommand{\descendants}[1]{\de{#1}}
\DeclareMathOperator{\lv}{lv}
\newcommand{\leaves}[1]{\lv{#1}}

\newcommand{\mvar}[1][X]{#1}
\newcommand{\mvars}[1][X]{\textbf{\textit{#1}}}
\newcommand{\mval}[1][x]{#1}
\newcommand{\mvals}[1][x]{\textbf{\textit{#1}}}
\newcommand{\valuespace}[1][\mvars]{\val{#1}}
\DeclareMathOperator{\val}{val}
\newcommand{\projection}[2]{#1\vert_{#2}}
\newcommand{\scope}[2][\mvars]{#1_{#2}}

\newcommand{\weight}[1][]{\textbf{\textit{w}}_{#1}}
\newcommand{\weightval}[2][\sumnode]{w_{\edge{#1}{#2}}}

\newcommand{\nf}[2][]{#2_{#1}} %
\newcommand{\nfval}[3][]{\nf[#1]{#2}(#3)} %

\begin{document}

\title{Soft Learning Probabilistic Circuits}

\author{Soroush Ghandi\inst{1} \and Benjamin Quost\inst{2} \and Cassio de Campos\inst{1}}

\institute{Eindhoven Unisversity of Technology \and University of Technology of Compiègne}

\maketitle              %

\begin{abstract}
Probabilistic Circuits (PCs) are prominent tractable probabilistic models, allowing for a range of exact inferences. This paper focuses on the main algorithm for training PCs, LearnSPN, a gold standard due to its efficiency, performance, and ease of use, in particular for tabular data. We show that LearnSPN is a greedy likelihood maximizer under mild assumptions. While inferences in PCs may use the entire circuit structure for processing queries, LearnSPN applies a \textit{hard} method for learning them, propagating at each sum node a data point through one and only one of the children/edges as in a hard clustering process. 
We propose a new learning procedure named \softl, that induces a PC using a \textit{soft} clustering process. We investigate the effect of this learning-inference compatibility in PCs. Our experiments show that \softl outperforms LearnSPN in many situations, yielding better likelihoods and arguably better samples. We also analyze comparable tractable models to highlight the differences between soft/hard learning and model querying. 

\end{abstract}

\section{Introduction}
Generative probabilistic models typically aim to learn the joint probability distribution of data, in order to perform probabilistic inference and answer queries of interest. However, not all the probabilistic models are the same in that regard. Models like variational autoencoders (VAEs) \cite{kingma2013auto} and generative adversarial networks (GANs) \cite{goodfellow2020generative} possess exceptional modeling prowess; nevertheless, their ability to perform probabilistic inference such as marginalization and conditioning is rather limited/intractable.

In contrast, \textit{tractable} probabilistic models, such as probabilistic circuits (PCs), including the prominent sum-product networks (SPNs)~ \cite{poon2011sum,sanchez2021sum}, allow for a wider range of exact inferences arguably at the expense of some power of fitness. PCs are a type of Probabilistic Graphical Models (PGMs), a class of models using a graph-based representation to encode high-dimensional distributions \cite{koller2009probabilistic}. Unlike Bayesian networks, which have a notoriously high complexity for general queries \cite{decampos2011new,decampos2005inferential}, PCs can produce several types of inferences in polynomial time under arguably mild assumptions \cite{vergari2021compositional}.

While intractable models such as VAE and GANs rely on deep neural networks as their structure, their PGM counterparts find graph structures that fit well the data. Moreover, PCs need to further reach structures that facilitate exact inferences, which translates into a constrained structure learning problem.
This latter problem 
has become an active line of research for many years. Many algorithms have been devised to learn PCs from data, among which LearnSPN \cite{gens2013learning} is considered a gold standard for its efficiency, performance, and ease of use. %
In addition to being the most widely known (and used) procedure for learning PCs---if not the best performing one in general, LearnSPN is also the building block of many subsequent algorithms \cite{lee2013online,molina2018mixed,vergari2015simplifying}. %

In a nutshell, LearnSPN follows a greedy search approach. The data is recursively partitioned into smaller chunks: the structure of the network is defined recursively, either by grouping variables (giving birth to product nodes) or clustering instances (resulting in sum nodes). We claim that this greedy learning approach may result in inappropriate clusters and lead to partitioning marginals rigidly at sub-optimal locations, which can potentially lead to overfitting and poor generalization. 

In this paper, we propose the \softl procedure as a counterpart to LearnSPN, with the aim to mitigate such potential drawbacks. \softl is a soft learning scheme akin to LearnSPN which may provide smoother marginals between data clusters so as to reduce the errors induced by misgrouped instances, and therefore lead to better likelihoods and arguably better samples. We compare \softl with LearnSPN over a range of datasets and configurations, and show that \softl manages to outperform LearnSPN in most cases, which empirically validates our claims regarding the potential improvements made by our soft learning scheme. We also draw comparisons with Cutset Networks~\cite{di2017fast}, another prominent tractable probabilistic model with similar goals.

\section{Related Work}

Arguably, the most common practical approach to learn the structure of PCs is to recursively partition the data matrix over the instances (forming sum nodes) and variables (forming product nodes) in a top-down fashion \cite{gens2013learning,molina2018mixed}. %
\cite{Adel2015} proposed to cluster the joint space of instances and variables, %
instead of alternating between instance and variable clustering. Conversely, another category of approaches consists in learning the structure in a bottom-up fashion by incrementally %
aggregating correlated variables \cite{hsu2017online,kalra2018online}. %

In another line of work, several attempts have been made to learn the structure of PCs in a more principled way, based on non-parametric formulations~\cite{trapp2016structure} %
and/or Bayesian structure learning~\cite{trapp2019bayesian,vergari2019automatic}. 
The Merged-L-SPN \cite{rahman2016merging} algorithm proposes to merge subtrees in a post-processing approach to reduce computation complexity and improve the generalization of LearnSPN. There have also been a variety of approaches aimed at learning other structures of tractable probabilistic models, such as probabilistic sentential decision diagrams (PSDDs) \cite{liang2017learning} and Cutset networks \cite{di2017fast,rahman2019look}.

\section{Probabilistic Circuits}

In this paper, we use the terms ``SPNs'' and ``PCs'' interchangeably. SPNs use circuits with three types of nodes: sum nodes can be interpreted as latent variables; product nodes encode context-specific independence; and leaves encode (tractable) univariate probability distributions~\cite{peharz_spn_latent_bn_2017}. %
Structurally, an SPN is a single-rooted directed acyclic graph (SDAG).
A directed graph is defined by a finite set $\nodes$ of nodes and a set $\edges\subseteq\nodes\times\nodes$ of ordered node pairs, called edges. 
For example, if $\node,\childnode\in\nodes$ and $(\node,\childnode)\in\edges$, then we have a directed edge between $\node$ and $\childnode$, written as $\edge{\node}{\childnode}$. 
A acyclic directed graph does not contain any directed cycle (i.e. a path of directed edges from any node to itself, non-directed cycles being allowed). 
We consider graphs with only one root (i.e., a node without incoming edge), denoted by $\rootnode$. 
We assume nodes can only belong to a single graph; thus, $\nodes$ and $\edges$ can be left implicit in the notation.

Any node $\node$ can be associated with important subsets: 
\begin{itemize}
	\item \emph{children} $\children{\node}\coloneqq\cset{\childnode\in\nodes}{\edge{\node}{\childnode}}$ are those nodes to which there is a directed edge from~$\node$,
	\item \emph{descendants} $\descendants{\node}\coloneqq\children{\node}\cup\bigcup_{\childnode\in\children{\node}}\descendants{\childnode}$ can be accessed by a sequence of directed edges from~$\node$,
	\item \emph{leaves} $\leaves{\node}\coloneqq\cset{\distrnode\in\descendants{\node}}{\children{\distrnode}=\emptyset}$ are descendants of~$\node$ that do not have children themselves.
\end{itemize}
Each node $\node$ of an SDAG determines a sub-SDAG $\subsdag{\node}$ rooted in~$\node$ with nodes $\set{\node}\cup\descendants{\node}$---the whole SDAG can be written as $\subsdag{\rootnode}$. Given a non-leaf node $\node$ of an SDAG, two of its children $\childnode_1, \childnode_2 \in \children{\node}$ are said to be either \emph{overlapping} if and only if $\leaves{\childnode_1} \cap \leaves{\childnode_2} \neq \emptyset$, or \emph{disjoint} if and only if $\leaves{\childnode_1} \cap \leaves{\childnode_2} = \emptyset$. 

Let $\mvars$ be a finite collection of random variables; $\mvar\in\mvars$ denotes that a random variable $\mvar$ belongs to $\mvars$ (collections of random variables and their realizations are in boldface, as opposed to single random variables and their realizations). 
Here, $\valuespace[\mvar]$ stands for the set of possible realizations~$\mval$ of $\mvar$, and $\valuespace$ for the set of possible realizations~$\mvals$ for the variables in $\mvars$, i.e. $\valuespace = \cartesianprod_{\mvar\in\mvars} \valuespace[X]$. 
Let $\mvars[Y]\subseteq \mvars$ be a subcollection of the variables in~$\mvars$: thus, if $\mvar\in\mvars[Y]$, then $\mvar\in\mvars$ (the converse being not necessarily true). 
Joint realizations $\mvals\in\valuespace$ or $\mvals[y]\in\valuespace[{\mvars[Y]}]$ can be projected onto a subspace.
For example, if $\mvars[Y]\subseteq \mvars$, then $\projection{\mvals}{\mvars[Y]}\in\valuespace[{\mvars[Y]}]$; we use the same notation $\projection{\mvals[y]}{\mvar}\in\valuespace[\mvar]$ to project $\mvals[y]$ onto a variable $\mvar\in\mvars[Y]$.

A SPN encodes a probabilistic model over a collection of variables $\mvars$ \cite{peharz2015theoretical,poon2011sum,sanchez2021sum}. 
It consists of an SDAG with structural constraints, and composed of three distinct types of nodes: sum nodes, associated with (numerical) parameters $\weight$; product nodes, and distribution nodes: these latter describe simple distributions at the leaves, which can be recursively combined using sums and products, so that the root encodes a complex distribution. 

Every node~$\node$ in a SPN is associated with a collection of random variables, called its \emph{scope}, over which it defines a probability distribution: e.g., $\scope{\node}$ stands for the scope of $\node$. 
The scope of a non-leaf node~$\node$ is the union of its child scopes: $\scope{\node}=\bigcup_{\childnode\in\children{\node}}\scope{\childnode}$. 
The root scope is $\scope{\rootnode}=\mvars$ (we assume every random variable to be in the scope of at least one leaf). 
We will write the projection of a node scope using the node symbol: $\projection{\mvals}{\node}\coloneqq\projection{\mvals}{\scope{\node}}$.

A \emph{sum node} $\sumnode$ in the SPN is associated with a function $\weight[\sumnode]$ on $\children{\sumnode}$ that returns edge weights $\weightval{\childnode}\coloneqq\weight[\sumnode](\childnode)$ for any $C\in\children{\sumnode}$: thus, $\weight[\sumnode]$ is the \emph{weight vector} associated with $\sumnode$. 
The weights should be non-negative, and we assume them normalized: for all $C\in\children{\sumnode}$, $\sum_{\childnode\in\children{\sumnode}}\weightval{\childnode}=1$ and $\weightval{\childnode}\geq 0$.
The numerical parameters $\weight$ of the entire SPN can simply be seen as the map $\weight:\sumnode \mapsto \weight[\sumnode]$ that identifies this vector for any sum node in the SPN.
The value $\nfval{\sumnode}{f}$ of any sum node $\sumnode$ is recursively obtained from the values of its children: 
    $\nfval{\sumnode}{f} \coloneqq \sum_{\childnode \in\children{\sumnode}} \weightval{\childnode}\nfval{\childnode}{f}$.
A \emph{product node} $\prodnode$ in the SPN %
is associated with a value $\nfval{\prodnode}{f}$, also recursively computed using the values of its children: 
    $\nfval{\prodnode}{f} \coloneqq \prod_{\childnode \in\children{\prodnode}} \nfval{\childnode}{f}$.
Finally, any leaf in the SDAG is a \emph{distribution node} $\distrnode$ in the SPN. Distribution nodes define a probability distribution over their scope $\scope{\distrnode}$. We will assume the scope $\scope{\distrnode}$ of any leaf node~$\distrnode$ to consist of a single random variable, and thus leaf nodes to encode univariate distributions. 

Assume we wish to calculate the expectation $\nfval{\rootnode}{f}\coloneqq\pexpswv{\rootnode}$ of a function $f$ according to the distribution encoded by the SPN with root~$\rootnode$ and weights $\weight$. 
Let $f$ be a product of indicators over variables in $\mvars$ (this allows for a range of probabilistic queries, including conditionals and marginals).
Let $\pexpm{\distrnode}$ denote the expectation operator with respect to the distribution in any leaf node $\distrnode$. 
The expectation $\nfval{\rootnode}{f}$ can be computed by propagating the expected values $\nfval{\distrnode}{f} \coloneqq \pexpmv[f_{\distrnode}]{\distrnode}$ in the leaves to the root node, thereby associating each node~$\node$ with an expected value $\nfval{\node}{f}$. %
Distribution nodes, being always the leaves in the SDAG, are the terminal points of the recursive definition of the node values. 
We assume in the sequel that we can \emph{efficiently evaluate} the expectations in the leaves, which opens the way to efficiently computing the expectation $\nfval{\rootnode}{f}$.

SPNs typically use additional structural assumptions to ensure the probabilistic model encoded is proper \cite{peharz2015theoretical}. For any sum node $\sumnode$ and any product node $\prodnode$ in the SPN, it holds that 
\begin{enumerate}[label=A\arabic*:, ref=A\arabic*]
    \item $\scope{\childnode_1}=\scope{\childnode_2}$ for all $\childnode_1,\childnode_2\in\children{\sumnode}$; \hfill \emph{(smoothness)}\label{assum1:smoothness}
    \item $\scope{\childnode_1}\cap\scope{\childnode_2}=\emptyset$ for all $\childnode_1,\childnode_2\in\children{\prodnode}$. \hfill \emph{(decomposability)}\label{assum2:decomp}
\end{enumerate}
A SPN that meets both \eqref{assum1:smoothness} and \eqref{assum2:decomp} is said to be \emph{valid}.

\section{Learning PCs} \label{sec:learning}

We first detail LearnSPN \cite{gens2013learning} and point out a potential drawback; then, we introduce our method \softl, explaining how it differs from LearnSPN, and illustrating how it might mitigate some issues.

LearnSPN employs a greedy search in the space of SPNs, and augments the network in a top-down fashion accordingly. It initializes the network with a single node $\rootnode$ representing the entire dataset (with scope $\mvars$), and then proceeds by recursively partitioning the dataset into smaller chunks based on instance/variable-wise groupings found in the data. For each variable-wise grouping, a product node $\prodnode$ is added to the network (representing a partition of the variables into conditionally independent groups); and each time instances are clustered, a sum node $\sumnode$ is added to the network (representing a mixture of the corresponding instances). This process is recursively applied until a stopping criterion is met, at which point each group of data corresponds to a univariate distribution that can be modeled reliably in the corresponding leaf of the network. 
Product nodes $\prodnode$ are created by using independence tests (pairwise tests will form a dependency graph, and variables in distinct components of the graph become the scope of the children of $\prodnode$), while sum nodes $\sumnode$ are created by performing hard clustering on the instances with the induced children having same scope as $\sumnode$. 

\begin{algorithm}[t]
    \caption{LearnSPN$(\projection{\instances}{\node}, \scope{\node})$ }%
    \label{alg:learnspn_correia}
    \begin{algorithmic}[1]
        \State \textbf{Input:} set of instances $\projection{\instances}{\node}\subseteq\valuespace[\scope{\node}]$ for a scope $\scope{N}$
        \State \textbf{Output:} an SPN $\subsdag{\node}$ representing a distribution over $\scope{\node}$ learned from $\projection{\instances}{\node}$
        \If{$|\scope{\node}| = 1$}
            \State $\node \leftarrow$ leaf univariate distribution node $\distrnode$ estimated from the variable's values in $\projection{\instances}{\node}$
        \Else
            \State partition $\scope{\node}$ into approximately independent subsets $\scope{\childnode_{j}}$, that is, $(\scope{\childnode_j})_{ j=1,\ldots,J}$ is a partition of $\scope{\node}$
            \If{$J>1$}\label{a1:success1}
                \State $\node\leftarrow$ $\bigotimes_{j=1}^J \text{LearnSPN}(\projection{\instances}{\childnode_j}, \scope{\childnode_j})$
            \Else
                \State partition $\projection{\instances}{\node}$ into subsets of similar instances $\projection{\instances_i}{\childnode_i}$, where $\childnode_i=\node$,  with $i=1,\dots,I$\label{a1:cluster}
                \If{$I>1$}
                    \State $\node\leftarrow$ $\bigoplus_{i=1}^I \frac{|\projection{\instances_i}{\childnode_i}|}{|\projection{\instances}{\node}|}\text{LearnSPN}(\projection{\instances_i}{\childnode_i}, \scope{\childnode_i})$
                \Else
                    \State $\node\leftarrow$ $\bigotimes_{j=1}^{|\scope{\node}|} \text{LearnSPN}(\projection{\instances}{\mvar_j}, \{\mvar_j\})$
                \EndIf
            \EndIf
        \EndIf\\
        \Return $\node$ 
    \end{algorithmic}
\end{algorithm}

The learning scheme of LearnSPN can seemingly be improved, as it appears not to be consistent with how queries in PCs work. During inference, a PC may use the entire structure to process a particular query, while the relevant information to that query is only used to train \emph{some parts} of the network structure, since at each sum node, a datapoint is propagated through one and only one of the children/edges as in a hard clustering process. Inherently, the response to a query will be mostly affected by the network parts trained on the relevant information; thus, this incompatibility does not lead to an erroneous response as long as the clustering in LearnSPN can well classify the queried datapoint. However, for datapoints that lie near the cluster borders, the response can become considerably erroneous in case of misgrouping by the clustering approach used in Line~\ref{a1:cluster} of Algorithm~\ref{alg:learnspn_correia}. 

Our proposal, \softl, induces a PC using a soft clustering process, so as to alleviate the costs of such misgrouping. Thus, after clustering, each datapoint is shared among the children of the sum node proportionally to its cluster memberships; this way, each datapoint will be propagated through the entire network, but with different weights indicating its importance to a particular part of the network. In order to do so, during clustering, for a set of $K$ clusters $\{ C_{1}, C_{2}, \dots, C_{K} \}$, a set of $K$ `datapoint' weights are associated to each datapoint, $\{ \dataw_{1}, \dataw_{2}, \dots, \dataw_{K} \}$ such that $\sum_{i} \dataw_{i} = 1$. Each instance in the dataset starts with a datapoint weight equal to 1 and this is divided among the children of sum nodes as the algorithm recursively performs clustering. 

However, propagating the weights associated with data points throughout the network and using them to learn a PC requires extensive adjustments to other parts of LearnSPN as well, beyond simply re-engineering the procedures, since clustering methods used to induce sum nodes, independence tests to induce product nodes, and distribution nodes at leafs need all to be adapted to deal with weighted datapoints.

We note that LearnSPN is a greedy maximizer of data likelihood, because each new node in the recursive construction can only increase the likelihood with respect to the alternative early stop of the procedure. This argument is detailed in Appendix~\ref{a4}. A number of adaptations and implementations of LearnSPN have been proposed. We use the version implemented by \cite{correia2020joints} (see Algorithm \ref{alg:learnspn_correia}) upon which we build our method, since such implementation achieved state of the art results. We nevertheless provide the results of the original implementation \cite{gens2013learning} in our experiments in Section \ref{exp}.

\subsection{Univariate density estimation} \label{ude}

Following \cite{correia2020joints}, we model the distribution over discrete or continuous variables using a multinomial or Gaussian distribution, respectively. These choices are not limiting, since PCs under this formulation can fit any distribution so long as structure learning is able to split the space properly and data are enough. In our case, each datapoint propagated in the learning algorithm has an associated weight. We employ the weights in our calculations as a measure of frequency. Take a univariate dataset $\projection{\instances}{\distrnode} = \{ d_{1}, d_{2}, \dots, d_{m} \}$ and a set of corresponding weights $\dataW = \{ \dataw_{1}, \dataw_{2}, \dots, \dataw_{m} \}$ at a particular leaf $\distrnode$, with $\dataw_i>0$ for all $i$ (datapoints with zero weight are discarded). For discrete variables, we will have:
\begin{equation}
    \widehat{P}(\scope{\distrnode} = k) = \dfrac{C_{k}}{\sum_{j=1}^mC_{j}} , \quad \text{ where } C_{j} = \sum_{i:~ d_{i} = j}\dataw_{i} .\label{eq:wdiscrete} 
\end{equation}
For continuous variables, we use a Gaussian $\gauss(\widehat{\mu},\widehat{\sigma})$ obtained with the proper derivations to achieve the reweighted estimation with Bessel's correction (derivations are omitted for ease of exposure): $\widehat{\mu} = (\sum_{i}\dataw_{i}d_{i})/(\sum_{i}\dataw_{i})$,
\begin{align}
\widehat{\sigma} & = \sqrt{\frac{\sum_{i=1}^m \dataw_i}{(\sum_{i=1}^m \dataw_i)^2 - \sum_{i=1}^m \dataw_i^2} \sum_{i=1}^m \dataw_i (d_i-\widehat{\mu})^2} . 
\end{align}

\subsection{Independence tests}

For discrete variables, \cite{correia2020joints} propose to use the chi-square test of independence, and to add a product node if the variables can be split into independent subsets. In brief, the chi-square test uses the difference between expected (under null hypothesis of independence) and observed ``frequencies'', using contingency tables.
In order to account for weighted instances, we form \textit{weighted contingency tables} and then proceed with the chi-square test based on the weighted tables. The calculations are similar to what is done in Expression~\eqref{eq:wdiscrete} (we leave for the reader to fill in the simple gaps to perform this computation).

In the case of two continuous variables or of mixed variables, we first discretize the continuous variables, and then apply the weighted chi-square test, while \cite{correia2020joints} use the Kruskal and Kendall's tau tests. Adapting Kruskal and Kendall's tau to deal with weighted data and meaningful commensurable interpretation among tests seems to be a challenge by itself. Our choice is motivated by the clear interpretation of partial frequencies that now all tests have in common, regardless of the data type, so all tests are commensurable. Their efficacy is not largely affected so long as enough data points are available (which is the case, as we would indeed want to stop expanding the model otherwise).

\subsection{Clustering}

In addition to handling soft membership degrees so as to quantify the relevance of an instance to a group, the clustering methods in our approach should also be able to admit weighted datapoints as inputs. We investigate two options. 
The first one is an adjusted version of the K-means clustering algorithm. In order to work with weighted data, we only change the update rule of centroids to account for datapoint weights. Originally, assuming a centroid $C$ is associated with $m$ datapoints $\{ d_{1}, d_{2}, \dots, d_{m} \}$, it would be updated as: 
$C_{\text{upd}} = (\sum_{i}d_{i})/m$.
In our adjusted version where datapoints are associated with weights $\{ \dataw_{1}, \dots, \dataw_{m} \}$, the update rule is simply
$C_{\text{new}} = {\sum_{i}\dataw_{i}d_{i}}/{\sum_{i}\dataw_{i}}$. 

Additionally, we utilize a weight function $\dataw(d, \{ C_{i} \}, i)$, that computes the weight $\dataw_{i}$ of each datapoint $d$ in each group $i$ based on the group centroid $C_{i}$. We define an arguably natural reweighted function as:
\begin{equation*}
    \dataw_{i} \coloneqq \dataw(d, \{ C_{i} \}, i) = \dfrac{\exp\{\beta(1 - \dfrac{||d - C_{i}||_{F}}{\sum_{i}||d - C_{i}||_{F}})\}}{\sum_{i}\exp\{\beta(1 - \dfrac{||d - C_{i}||_{F}}{\sum_{i}||d - C_{i}||_{F}})\}} . 
\end{equation*}
To put it in words, this reweighted function (i) computes the distances of the datapoint $d$ to group centroids $\{ C_{i} \}_{\forall i}$ and normalizes them; (ii) computes an intermediate relevancy degree $1 - ||d - C_{i}||_{F}/\sum_{i}||d - C_{i}||_{F}$ to each group, which trivially gives more (resp. less) value to groups that are closer to (resp. further away from) the datapoint; and (iii) applies a softmax function to the relevancy degrees.

The second clustering method we investigate is based on estimating mixtures of distributions using the Expectation-Maximization (EM) algorithm, under a conditional independence assumption. Let the dataset be $\projection{\instances}{\sumnode}$ over variables $\projection{\mvars}{\sumnode}$. In order to perform EM clustering, we assume the underlying distribution to be a mixture of $c$ fully factorized distributions ($c$ is the targeted number of children for $\sumnode$): 
$P(\mvars) = \sum_{i = 1}^{c} P(i)\prod_{\mvar\in\mvars} P(\mvar|i)$. 
In a nutshell, EM estimates the model parameters by iteratively alternating between an expectation and a maximization step, until a stopping criterion is met. The expectation step amounts to updating the (soft) memberships of the instances to the clusters, based on the current distribution estimates; the maximization step, to update the univariate distributions $P(\mvar|i)$ and the group priors $P(i)$ based on the new memberships. %
In order to make EM work with weighted data, we only need to make two changes to the algorithm: we modify the univariate distribution updates in the maximization step to include the weights, as for univariate distribution estimation in Section~\ref{ude}, %
and we make the updates for the group priors $P(i)$ proportional to the sum of weights (instead of whole counts). Both steps are repeated iteratively until convergence to a stationary point (as usual in EM).
The adjustments result in a soft learning scheme \softl, whose pseudocode is given in Algorithm~\ref{alg:learnspn_soft}, which is not only more compatible with the soft inference scheme of PCs, but also uses each datapoint to learn every part of the network. 

\begin{algorithm}[t]
    \caption{$\softl(\projection{\instances}{\node}, \scope{\node}, \projection{\dataW}{\node})$ }
    \label{alg:learnspn_soft}
    \begin{algorithmic}[1]
        \State \textbf{Input:} set of instances $\projection{\instances}{\node}\subseteq\valuespace[\scope{\node}]$ for a scope $\scope{N}$, set of weights $\projection{\dataW}{\node}$ corresponding to datapoint instances
        \State \textbf{Output:} an SPN $\subsdag{\node}$ representing a distribution over $\scope{\node}$ learned from $\projection{\instances}{\node}$
        \If{$|\scope{\node}| = 1$}
            \State $\node \leftarrow$ leaf univariate distribution node $\distrnode$ estimated from the variable's values in $\projection{\instances}{\node}$ with weights in $\projection{\dataW}{\node}$
        \Else
            \State partition $\scope{\node}$ into approximately independent subsets $\scope{\childnode_{j}}$ using weighted independence tests with weights $\projection{\dataW}{\node}$, that is, $(\scope{\childnode_j})_{ j=1,\ldots,J}$ is a partition of $\scope{\node}$, using $\projection{\instances}{\node}$ and $\projection{\dataW}{\node}$
            \If{$J>1$}\label{a2:success1}
                \State $\node\leftarrow$ $\bigotimes_{j=1}^J \softl(\projection{\instances}{\childnode_j}, \scope{\childnode_j},\projection{\dataW}{\childnode_j})$
            \Else
                \State partition $\projection{\instances}{\node}$ using a weighted soft clustering with datapoint weights $\projection{\dataW}{\node}$, yielding new weights $\{ \dataW_{i} \}_{\forall i}$, with $i=1,\dots,I$ ($I$ is the number of groups)\label{a2:cluster}
                \State update $\dataW_{i}\leftarrow \dataW_{i} \cdot \projection{\dataW}{\node}$ and let $s_i=\sum\dataW_{i}$                %
                \If{$I>1$}
                    \State $\node\leftarrow$ $\bigoplus_{i=1}^I \frac{s_i}{\sum_j s_j} \softl(\projection{\instances}{\childnode_i}, \scope{\childnode_i}, \dataW_{i})$
                \Else
                    \State $\node\leftarrow$ $\bigotimes_{j=1}^{|\scope{\node}|} \softl(\projection{\instances}{\mvar_j}, \{\mvar_j\}, \projection{\dataW}{\mvar_j})$
                \EndIf
            \EndIf
        \EndIf\\
        \Return $\node$ 
    \end{algorithmic}
\end{algorithm}

\begin{figure}[t]
    \centering
    \resizebox{0.5\textwidth}{!}{\definecolor{blu}{RGB}{199, 206, 234}
\definecolor{gre}{RGB}{181, 234, 215}
\definecolor{re}{RGB}{255, 154, 162}
\definecolor{ore}{RGB}{255, 218, 193}
\definecolor{lgr}{RGB}{226, 240, 203}
\definecolor{mel}{RGB}{255, 183, 178}

\begin{tikzpicture}[
    cross/.style={minimum size=6mm, fill=lgr,
        path picture={
            \draw[black] (path picture bounding box.south east) -- (path picture bounding box.north west) (path picture bounding box.south west) -- (path picture bounding box.north east);
        }
    },
    sum/.style={minimum size=6mm, fill=blu,
        path picture={
            \draw[black] (path picture bounding box.south) -- (path picture bounding box.north) (path picture bounding box.west) -- (path picture bounding box.east);
        }
    },
    integral/.style={minimum size=6mm, fill=re, 
        inner sep=0pt,
        text width=4mm,
        align=center,},
    gauss/.style={minimum size=6mm, fill=ore,
        path picture={
            \draw[black] plot[domain=-.15:.15] ({\x},{exp(-200*\x*\x -2.)});
        }
    },
    indicator/.style={minimum size=6mm, fill=ore,
        path picture={
            \draw[black] plot[domain=-.1:.1] ({\x},{((\x > 0) - 0.5)*0.35});
        }
    },
    dt/.style={minimum size=5mm, fill=white
    },
    line/.style={
      draw, thick,
      -latex',
      shorten >=2pt
    },
    cloud/.style={
      draw=red,
      thick,
      ellipse,
      fill=red!20,
      minimum height=1em
    },
    spn/.style={
    regular polygon,
    regular polygon sides=3,
    draw,
    fill=white
    }
]

    \node[draw, circle, sum] (s1) at (0, 0) {};
    \node[draw, circle, cross] (p1) at (-2, -1) {};
    \node[draw, circle, cross] (p2) at (2, -1) {};
    \node[draw, circle, gauss] (g1) at (-3, -2.5) {};
    \node (b) at (-3, -3) {\footnotesize $\mathcal{N}_{X}(-0.5, 1)$};
    \node (b) at (-1, -3) {\footnotesize $\mathcal{N}_{Y}(-2, 0.2)$};
    \node (b) at (1, -3) {\footnotesize $\mathcal{N}_{X}(0.5, 1)$};
    \node (b) at (3, -3) {\footnotesize $\mathcal{N}_{Y}(2, 0.2)$};
    \node[draw, circle, gauss] (g2) at (-1, -2.5) {};
    \node[draw, circle, gauss] (g3) at (1, -2.5) {};
    \node[draw, circle, gauss] (g4) at (3, -2.5) {};

    \draw[line width=0.3mm, ->] (s1) to node[black, yshift=0.25cm]{$0.5$} (p1);
    \draw[line width=0.3mm, ->] (s1) to node[black, yshift=0.25cm]{$0.5$} (p2);
    \draw[line width=0.3mm, ->] (p1) to (g1);
    \draw[line width=0.3mm, ->] (p1) to (g2);
    \draw[line width=0.3mm, ->] (p2) to (g3);
    \draw[line width=0.3mm, ->] (p2) to (g4);

\end{tikzpicture}}
    \caption{PC structure equivalent to Expression~\eqref{eq:genbadclus}, with a root sum node (in blue) with balanced weights to its children, which are two product nodes (in green), and four leaf distribution nodes (in salmon).}
    \label{fig:SPN-example}
\end{figure}
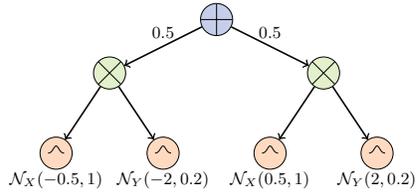

\begin{figure}[t]
    \centering
        \includegraphics[width=0.6\textwidth]{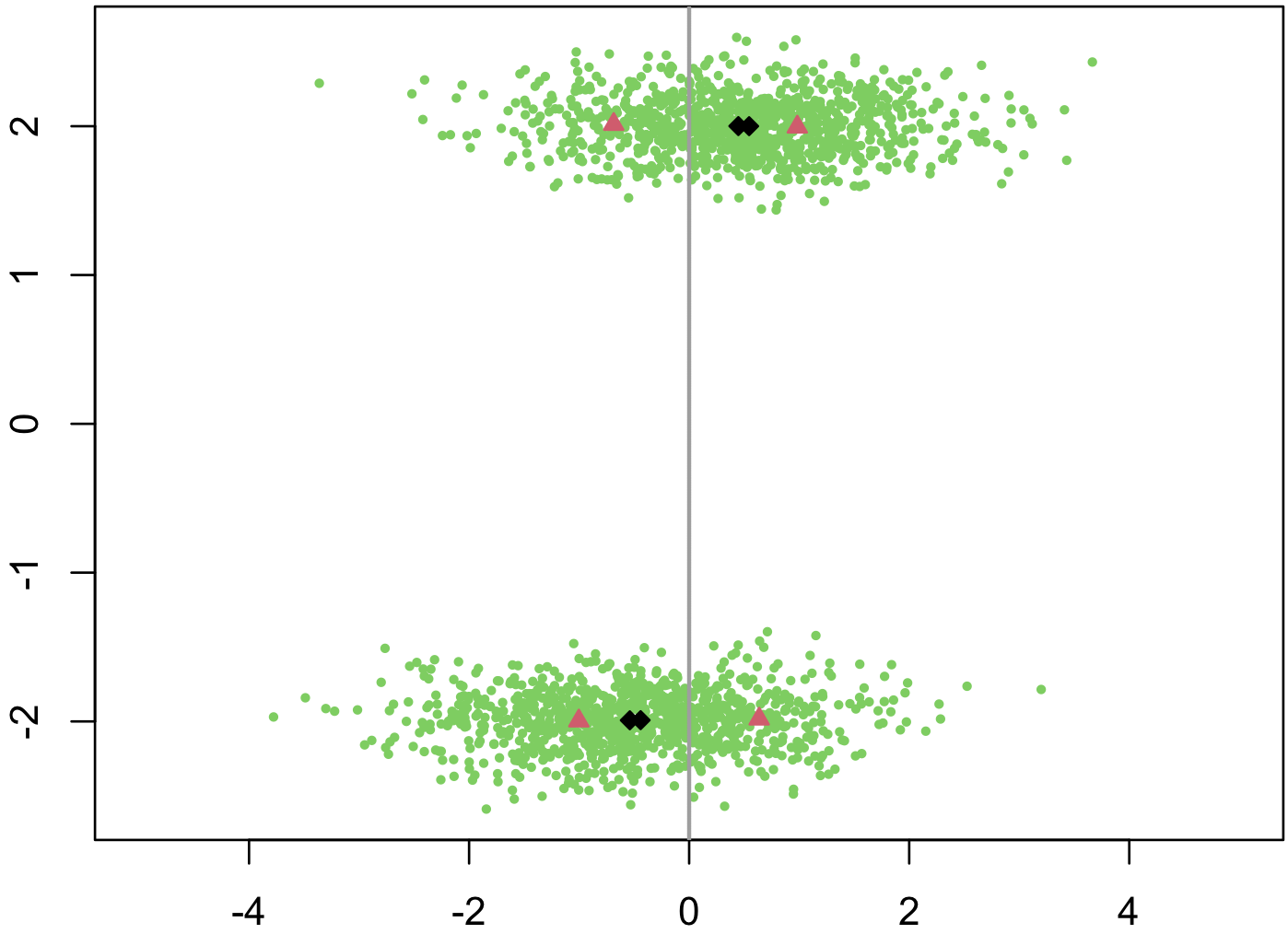}
    \hfill
    \caption{Green points $(X,Y)$ (resp. horizontal and vertical axes as usual) generated from the PC in Expression~\eqref{eq:genbadclus}. The gray line is a hypothetical bad partition obtained for the root node. %
    \softl yields the mean parameters of the Gaussian leafs represented by the black diamonds, which still captures the whole Gaussians on both sides of the gray cut of the first step; standard LearnSPN yields the red triangles as means, as both clusters are necessarily treated separately.}
    \label{fig:badcluster}
\end{figure}

We close this section with an (extreme-case) illustrative example of the potential benefits of \softl (so please bear with us). Assume that data $(X,Y) \in \realns^2$ is generated from a PC as per Equation~\eqref{eq:genbadclus} (we use a flat notation for the PC): 
\begin{align}
(X,Y)\sim~ &0.5\cdot \gauss_X(-0.5,1)\otimes \gauss_Y(-2,0.2) \oplus
0.5\cdot\gauss_X(0.5,1)\otimes \gauss_Y(2,0.2).\label{eq:genbadclus}
\end{align}
Figure \ref{fig:SPN-example} shows the equivalent graphical representation, which respects Assumptions \ref{assum1:smoothness}--\ref{assum2:decomp} and hence induces a valid joint distribution for $X,Y$. Figure~\ref{fig:badcluster} shows a sample with 1000 datapoints (green points) for each of the two mixture components, which are independent bivariate Gaussians. Sampling is performed top-down, choosing the direction to take in a sum node based on its weights, while following all paths from product nodes, yielding a full sample when the corresponding leaf nodes are reached and sampled. This generating distribution is obviously unknown to the learning algorithms.

We start running both LearnSPN and \softl on this dataset. 
Initially, the conditions in Line~\ref{a1:success1} of Algorithm~\ref{alg:learnspn_correia} and in Line~\ref{a2:success1} of Algorithm~\ref{alg:learnspn_soft} both fail, $X$ and $Y$ being not independent ($X$ is shifted based on $Y$), and the algorithms proceed with creating a sum node. 
Assume now that the clustering fails and partitions the points as per the gray line in Figure~\ref{fig:badcluster} (that is, at $X=0$). Hence, in the continuation, LearnSPN will have to recursively deal with the groups of points in the left- and right-side of the gray line separately and independently, while \softl will weight the pertinence of each data point to each of the two groups. 
After that first sum node, a product node will not appear again ($X$ and $Y$ being still considerably dependent). Assume now that the next clusterings run to create new sum nodes work perfectly well (otherwise, the difference in favor of \softl could be even stronger, as we will see), and hence split points perfectly (positive $Y$ go to one side and negative $Y$ to the other). Finally, $X$ and $Y$ will be found independent (enough) and the four bivariate Gaussians will appear as in the following expressions.
LearnSPN gives the model in Expression~\eqref{eq:exlspn}: $(X,Y)\sim$
\begin{align}
\sim .49 (& 0.7 \cdot\gauss_X( -1 , .69 )\otimes \gauss_Y( -2 , .2 ) \oplus
 .3 \cdot\gauss_X( -.69 , .54 )\otimes \gauss_Y( 2.01 , .2 )) \oplus\nonumber\\
.51 (& .31 \cdot\gauss_X( .64 , .53 )\otimes \gauss_Y( -1.98 , .2 ) \oplus
 .69 \cdot\gauss_X( .98 , .68 )\otimes \gauss_Y( 2 , .19 )) ; 
\label{eq:exlspn}
\end{align}
\noindent and \softl the model in Expression~\eqref{eq:exsoftl}: $(X,Y)\sim$
\begin{align}
\sim
.5 (& .5 \cdot\gauss_X( -.51 , 1.01 )\otimes \gauss_Y( -1.99 , .2 ) \oplus
 .5 \cdot\gauss_X( .48 , .99 )\otimes \gauss_Y( 2 , .2 )) \oplus\nonumber\\
.5 (& .51 \cdot\gauss_X( -.47 , .99 )\otimes \gauss_Y( -1.99 , .2 ) \oplus
 .49 \cdot\gauss_X( .52 , 1 )\otimes \gauss_Y( 2 , .2 )) . 
\label{eq:exsoftl}
\end{align}

In this hypothetical example, both approaches learn reasonable parameter estimates for the distribution leafs over $Y$ (the true generating distribution is in Expression~\eqref{eq:genbadclus}), but LearnSPN struggles to get good estimates for $X$ due to the bad clustering at the first sum node (the true should be Gaussians $\gauss(0.5,1))$ and $\gauss(-0.5,1))$). 
On the other hand, \softl has less difficulty to achieve good estimates for $X$. The Gaussian means of the distribution leaf nodes obtained by LearnSPN are shown in Figure~\ref{fig:badcluster} as red triangles, while the same is shown for \softl in black diamonds. The figure clearly shows the difference between the hard LearnSPN approach and \softl, suggesting that the latter might better cope with bad clustering results. Arguably, such results are inevitable when performing clustering with high-dimensional heterogeneous data---even though they are unlikely to be in practice as bad as in this example. It is not hard to imagine node pruning/merging techniques that would make the outcome of \softl more compact in this situation, as some terms in Expression~\eqref{eq:exsoftl} represent somewhat similar Gaussians (and by that potentially we could recover even the simpler true structure of Expression~\eqref{eq:genbadclus}).

\section{Experiments} \label{exp}

We conduct a variety of experiments to evaluate our hypothesis about learning-inference compatibility and potential drawbacks of LearnSPN, by comparing its performance against our proposal \softl. We proceed in three steps: (i) we compare the test log-likelihood of \softl against that of LearnSPN on a variety of discrete and mixed datasets; (ii) we visually compare the quality of samples generated by \softl and LearnSPN on an image dataset; and (iii) we numerically compare the quality of samples generated by \softl and LearnSPN on discrete datasets.

On discrete (binary) datasets, \softl is compared against both implementations of LearnSPN \cite{correia2020joints,gens2013learning}. For the rest of the experiments, since we have no access to the original implementation of LearnSPN \cite{gens2013learning}, the comparison is held between \softl and LearnSPN \cite{correia2020joints}. Throughout the experiments, both algorithms are optimized over two sets of hyperparameters, namely the chi-square test significance $p \in \{0.01, 0.001, 0.0001 \}$ and the Laplace smoothing parameter of multinomial density estimation $\alpha \in \{ 0.1, 0.01, 10^{-6} \}$ (as usual on discrete data counts). %

\subsection{Test log-likelihood} \label{tll}

\subsubsection{Binary datasets.}
We compare the test log-likelihood of our method against LearnSPN \cite{correia2020joints,gens2013learning} on twenty real-world datasets, of which thirteen were introduced in \cite{lowd2010learning}, and the other seven in \cite{van2012markov}. The number of instances in the datasets varies from 2K to 388K, and the number of variables from 16 to 1556. In addition, we include the results of CNET \cite{rahman2014cutset}, as this latter method combines a hard learning scheme with a hard inference scheme, representing the other side of the learning-inference spectrum. Note that \cite{rahman2014cutset} report 3 sets of results for 3 different versions of CNET, among which MCNET (which consists of an ensemble of CNETs) shows strong results and outperforms our method on several datasets; however, since our method does not include an ensemble of models and/or pruning, we only report the results for the vanilla CNET, as it is the most comparable version to \softl (we considered beyond the scope to compare ensemble methods). %
The results for LearnSPN \cite{gens2013learning} and CNET \cite{rahman2014cutset} are reported from their corresponding paper. The results for LearnSPN \cite{correia2020joints} and \softl are average of 9 repetitions (random initializations), and are summarized in Table \ref{tab:res_bin_gens}.

\begin{table}[t]
    \centering
    \small
    \begin{tabular}{|l|c|c|c|c|}
        \toprule
        \multirow{2}{*}{\textbf{Data}} & \multicolumn{2}{c|}{\textbf{LearnSPN}} & \multirow{2}{*}{\textbf{CNET}} & \textbf{Soft} \\
        \cline{2-3}
         & \textbf{Gens} & \textbf{Correia} &  & \textbf{Learn} \\
        \hline
         NLTCS & -6.11 & -5.99 & -6.10 & \textbf{-5.97} \\
         MSNBC & -6.11 & \textbf{-6.04} & -6.06 & -6.04 \\
         KDD-2k & \textbf{-2.18} & -2.35 & -2.21 & -2.34 \\
         Plants & -12.97 & -12.87 & -13.37 & \textbf{-12.57} \\
         Audio & -40.50 & -39.84 & -46.84 & \textbf{-39.65} \\
         Jester & -75.98 & -53.23 & -64.50 & \textbf{-53.00} \\
         Netflix & -57.32 & -56.82 & -69.74 & \textbf{-56.49} \\
         Accid. & -30.03 & \textbf{-28.89} & -31.59 & -29.54 \\
         Retail & -11.04 & -11.09 & -11.12 & \textbf{-10.88} \\
         Pumsb. & -24.78 & \textbf{-24.10} & -25.06 & -24.81 \\
         \bottomrule
    \end{tabular}
    \begin{tabular}{|l|c|c|c|c|}
        \toprule
        \multirow{2}{*}{\textbf{Data}} & \multicolumn{2}{c|}{\textbf{LearnSPN}} & \multirow{2}{*}{\textbf{CNET}} & \textbf{Soft} \\
        \cline{2-3}
         & \textbf{Gens} & \textbf{Correia} &  & \textbf{Learn} \\
        \hline
         DNA & -82.52 & -83.67 & -109.79 & \textbf{-82.06} \\
         Kosarek & -10.98 & -11.04 & -11.53 & \textbf{-10.89} \\
         MSWeb & -10.25 & -9.85 & -10.20 & \textbf{-9.68} \\
         Book & -35.88 & -34.33 & -40.19 & \textbf{-33.03} \\
         E.Movie & \textbf{-52.48} & -56.84 & -60.22 & -55.22 \\
         WebKB & \textbf{-158.20} & -159.53 & -171.95 & -158.70 \\
         Reut.52 & \textbf{-85.06} & -87.93 & -91.35 & -88.33 \\
         20ng  & -155.92 & -122.16 & -176.56 & \textbf{-121.09} \\
         BBC & -250.68 & \textbf{-247.81} & -300.33 & -249.38  \\
         Ad & -19.73 & -18.53 & \textbf{-16.31} & -20.30  \\
         \bottomrule
    \end{tabular}
    \caption{Performance results of \softl vs. LearnSPN \cite{gens2013learning}, LearnSPN \cite{correia2020joints}, and CNET \cite{rahman2014cutset} over binary datasets.}
    \label{tab:res_bin_gens}
\end{table}

\subsubsection{Mixed datasets.}
We compare the test log-likelihood of the methods over a selection of datasets from the OpenML-CC18 benchmark \cite{vanschoren2014openml}. Table \ref{tab:res_mix} presents the results averaged over 9 repetitions of the algorithm.

\begin{table}[t]
    \centering
    \small
    \begin{tabular}{|l|c|c|c|}
        \toprule
        \textbf{Dataset} & \textbf{LearnSPN} & \textbf{\softl} & \textbf{Early \softl} \\
        \hline
        bank & -20.139 & \textbf{-19.993} & \textbf{-19.997} \\
        electricity & -11.229 & \textbf{-11.217} & -11.226  \\
        segment & -17.517 & \textbf{-17.480} & -17.493 \\
        german & -22.720 & \textbf{-22.395} & -22.470\\
        vowel & -16.957 & \textbf{-16.584} & -16.634 \\
        cmc & -9.850 & \textbf{-9.811} & -9.838\\
         \bottomrule
    \end{tabular}
    \caption{Performance results of LearnSPN vs. \softl on mixed datasets. Last column uses \softl with at most 2 iterations within clustering algorithms, thus forcing their early stop (often before convergence).}
    \label{tab:res_mix}
\end{table}

As the results in Table \ref{tab:res_bin_gens} suggest, \softl manages to outperform both implementations of LearnSPN on 14 out of 20 discrete datasets, and to outperform CNET on 18 out of 20 datasets. It also performs better than LearnSPN \cite{correia2020joints} on all mixed datasets. This indicates that a soft learning scheme can have a positive impact on the performance of LearnSPN over discrete and mixed datasets. We attribute this to learning-inference compatibility caused by the soft learning scheme. \softl results in softer margins between groups when clustering: as a result, if a datapoint is misgrouped, the induced error will not be as costly as with the original LearnSPN. We would also like to remind that for mixed datasets in \softl, continuous variables are discretized for each independence test, which adds another layer of estimation to the algorithm; however, \softl still manages to outperform its counterpart.

We also empirically study the possibility of early stopping clustering algorithms to try to demonstrate the greater robustness of \softl to potentially worse clustering results (with the benefit of speeding up the learning, as one can have fewer iterations). We limited clustering to only 2 iterations (often stopping before convergence of the method). The results (last column in Table~\ref{tab:res_mix}) suggest that the speed up comes with little harm to the accuracy of the model, which still outperforms LearnSPN run without early stopping (even though we acknowledge that broader experiments are needed for a more conclusive claim in this regard).

\subsection{Image data}

\begin{figure}[t]
    \centering
    \begin{subfigure}[b]{0.33\textwidth}
        \centering
        \includegraphics[width=\textwidth]{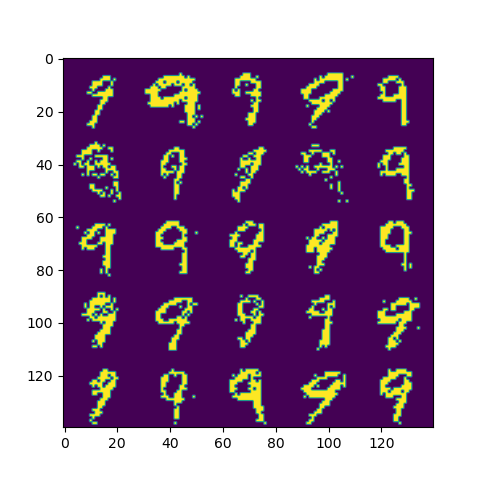}
    \end{subfigure}
    \begin{subfigure}[b]{0.33\textwidth}
        \centering
        \includegraphics[width=\textwidth]{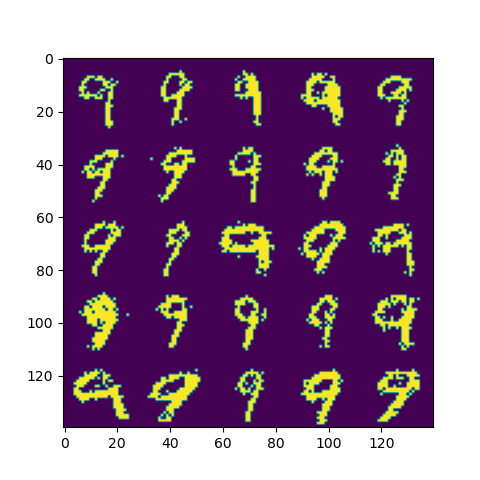}
    \end{subfigure}
    \begin{subfigure}[b]{0.33\textwidth}
        \centering
        \includegraphics[width=\textwidth]{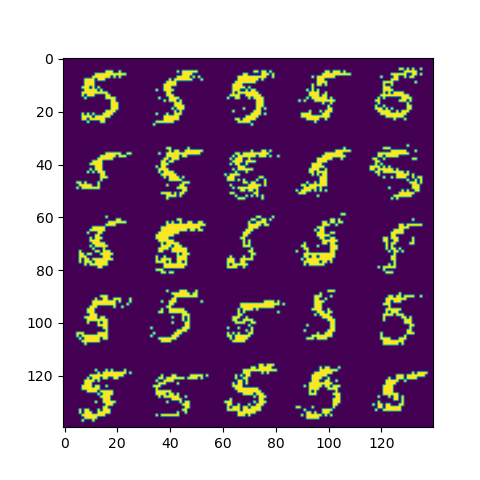}
    \end{subfigure}
    \begin{subfigure}[b]{0.33\textwidth}
        \centering
        \includegraphics[width=\textwidth]{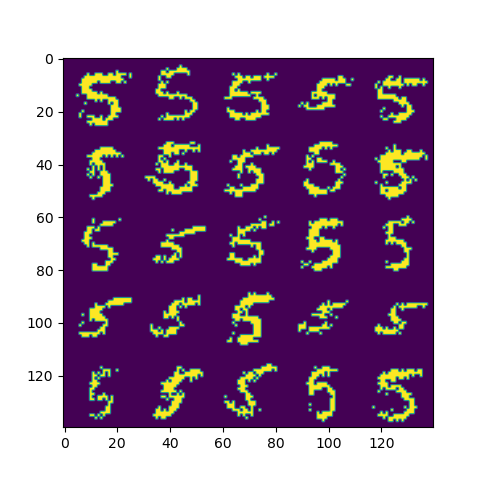}
    \end{subfigure}
    \caption{Samples from PCs trained on Binary MNIST (n.9 and n.5), using (left) LearnSPN and (right) \softl.}
    \label{fig:bin-mnist}
\end{figure}

We employ LearnSPN \cite{correia2020joints} and \softl to learn PCs over the binary-MNIST \cite{larochelle2011neural} dataset, and then we qualitatively evaluate the generated samples from both PCs. We decide to learn each PC on a single class of the dataset at a time in order to better visualize the samples. Figure \ref{fig:bin-mnist} shows the generated samples from PCs learned on classes 9 and 5 of the dataset, respectively. As the results would suggest, the samples generated from the PC learned by \softl appear to be less cluttered. We believe the soft clustering to be responsible for some better samples when compared to those generated from the PC trained with LearnSPN. Samples drawn from \softl are mostly clear to interpret, even if they can also be noisy or erroneous.

\subsection{Quality of generated samples}

Here we aim to evaluate and compare the quality of samples generated from PCs learned via LearnSPN and \softl. To do so, we follow the experiment setup proposed in \cite{fakoor2020trade}. For each algorithm and dataset, we (i) learn a PC over the training dataset (using the best-performing set of hyperparameters); (ii) generate a synthetic dataset using the learned PC; (iii) learn another PC over the generated synthetic dataset; and (iv) interpret the test log-likelihood of learned PCs as an indicator for the quality of generated samples. The results of this experiment over 5 datasets are summarized in Table \ref{tab:QOS}.
While \softl still manages to outperform LearnSPN on 4 out of 5 synthetic datasets, it shows a greater performance drop in 3 out of 5 experiments. This shows that while \softl objectively generates better samples, whether or not its performance is more affected by the synthetic data remains inconclusive.

\begin{table}[t]
    \centering
    \small
    \begin{tabular}{|l|c|c|c|c|}
        \toprule
        \multirow{3}{*}{\textbf{Dataset}}  & \multicolumn{4}{c|}{\textbf{Test Log Likelihood}} \\
        \cline{2-5}
         & \multicolumn{2}{c|}{\textbf{LearnSPN}} & \multicolumn{2}{c|}{\textbf{\softl}} \\
         \cline{2-5}
         & \textbf{Original} & \textbf{Synthetic} & \textbf{Original} & \textbf{Synthetic} \\
        \hline
         NLTCS & -5.997 & -6.051 & -5.976 & -6.022 \\
         Audio & -39.823 & -40.307 & -39.649 & -40.143 \\
         Retail & -11.074 & -11.196 &-10.880 & -10.974 \\
         MSWeb & -9.833 & -10.058 &-9.696 & -9.982 \\
         Reuters-52 & -87.838 & -90.741 &-88.609  & -97.002\\
         \bottomrule
    \end{tabular}
    \caption{Performance drop for PCs trained on synthetically generated samples, averaged over 3 repetitions.}
    \label{tab:QOS}
\end{table}

\section{Conclusion}

In this paper, we shed some light on the importance of learning-inference compatibility of PCs and the potential drawbacks of greedy algorithms such as LearnSPN, which can potentially lead to rigid partitions and poor generalization. We also introduced \softl, a soft structure-learning scheme as an attempt to mitigate the costs of such greedy behaviors. Our experiments showed that this soft method outperforms LearnSPN on a variety of datasets and configurations on test likelihoods, and that it arguably generates better samples likely due to its smoother partition margins. 

This paper attempts to push a reasonably simple idea of soft clustering, yet with intricate changes required in the clustering and independence test methods. We truly believe structure learning to be a major point of improvement for PCs to reach even greater accuracy in real-world applications, in particular for structured/tabular data. Multiple avenues remain to be explored in learning the structure of PCs, which constitute excellent future work. We intend to continue the study with pruning/merging techniques and to move away from excessively structure-learning greedy approaches. In some sense, \softl is a partial step in that direction by trying to mitigate greediness.

\begin{credits}

\subsubsection{\discintname}
The authors have no competing interests to declare that are relevant to the content of this article.
\end{credits}
\bibliographystyle{splncs04}
\bibliography{refs}

\newpage
\appendix
\section{Dataset Details}

\begin{table}[H]
    \centering
    \small
    \begin{tabular}{|l|c|c|c|c|c|}
        \toprule
         \textbf{Dataset} & \textbf{Vars.} & \textbf{Train} & \textbf{Valid.} & \textbf{Test} & \textbf{Density} \\
         \hline
         NLTCS & 16 & 16181 & 2157 & 3236 & 0.332 \\
         MSNBC & 17 & 291326 & 38843 & 58265 & 0.166 \\
         KDDCup2k & 65 & 180092 & 19907 & 34955 & 0.008 \\
         Plants & 69 & 17412 & 2321 & 3482 & 0.180 \\
         Audio & 100 & 15000 & 2000 & 3000 & 0.199 \\
         Jester & 100 & 9000 & 1000 & 4116 & 0.608 \\
         Netflix & 100 & 15000 & 2000 & 3000 & 0.541 \\
         Accidents & 111 & 12758 & 1700 & 2551 & 0.291 \\
         Retail & 135 & 22041 & 2938 & 4408 & 0.024 \\
         Pumsb-star & 163 & 12262 & 1635 & 2452 & 0.270 \\
         DNA & 180 & 1600 & 400 & 1186 & 0.253 \\
         Kosarak & 190 & 33375 & 4450 & 6675 & 0.020 \\
         MSWeb & 294 & 29441 & 3270 & 5000 & 0.010 \\
         Book & 500 & 8700 & 1159 & 1739 & 0.016 \\
         EachMovie & 500 & 4524 & 1002 & 591 & 0.059 \\
         WebKB & 839 & 2803 & 558 & 838 & 0.064 \\
         Reuters-52 & 889 & 6532 & 1028 & 1540 & 0.036 \\
         20 Newsgrp. & 910 & 11293 & 3764 & 3764 & 0.049 \\
         BBC & 1058 & 1670 & 225 & 330 & 0.078 \\
         Ad & 1556 & 2461 & 327 & 491 & 0.008 \\
         \bottomrule
    \end{tabular}
    \caption{Discrete datasets statistics.}
    \label{tab:bindata_stats}
\end{table}

\begin{table}[H]
    \centering
    \small
    \begin{tabular}{|l|c|c|c|}
    \toprule
         \multirow{2}{*}{\textbf{Dataset}} & \multicolumn{2}{c|}{\textbf{Vars.}} & \multirow{2}{*}{\textbf{Size}} \\
         \cline{2-3}
         & \textbf{Categorical} & \textbf{Numeric} &  \\
         \hline
         bank & 10 & 7 & 45211 \\
         electricity & 2 & 7 & 45312 \\
         segment & 1 & 19 & 2310 \\
         german & 14 & 7 & 1000 \\
         vowel & 3 & 10 & 990 \\
         cmc & 8 & 2 & 1473 \\

    \bottomrule
    \end{tabular}
    \caption{Mixed datasets statistics.}
    \label{tab:mixdata_stats}
\end{table}

As mentioned in Section \ref{tll}, we use a set of twenty real-world binary datasets \cite{lowd2010learning,van2012markov} and 6 real-world mixed datasets \cite{vanschoren2014openml} for the log-likelihood experiments. Over the binary datasets, the number of instances varies from 2K to 388K, and the number of variables from 16 to 1556; over the mixed datasets, the number of variables varies from 9 to 21, and the number of instances from 990 to 45312. The details regarding these datasets are outlined in Tables \ref{tab:bindata_stats} and \ref{tab:mixdata_stats}. Note that the binary datasets are already divided into train, validation, and test sets, and hence, the details regarding the size of each set are also reported in the table.

\section{Experiment Details}

In this section, we present supplementary results and details of our experiments, in addition to the results reported in Section \ref{exp}. For each method (LearnSPN and \softl), and for each dataset category (binary and mixed), we present a table outlining more details regarding the experiments. Each table provides a more detailed version of the results, which decomposes the results over the clustering method and provides the standard deviation of the results over the 9 repetitions of the experiments. In addition to that, each table also provides the hyperparameter configurations to reproduce the reported results. Tables \ref{tab:lspn_bin_det} and \ref{tab:softl_bin_det} are dedicated to the experiments of LearnSPN and \softl on binary datasets. Similarly, tables \ref{tab:lspn_mix_det} and \ref{tab:softl_mix_det} are dedicated to respective experiments of LearnSPN and \softl on mixed datasets.

\begin{table}[H]
    \centering
    \small
    \begin{tabular}{|l|c|c|c|c|c|c|}
        \toprule
        \multirow{2}{*}{\textbf{Dataset}} & \multicolumn{6}{c|}{\textbf{LearnSPN}} \\
        \cline{2-7}
         & \multicolumn{3}{c|}{\textbf{EM}} & \multicolumn{3}{c|}{\textbf{K-means}} \\
         \cline{2-7}
         & \textbf{Test LL} & \textbf{$p$} & \textbf{$\alpha$} & \textbf{Test LL} & \textbf{$p$} & \textbf{$\alpha$} \\
        \hline
         NLTCS & -5.997 $\pm$ 0.008 & 0.01 & 0.01 & -5.995 $\pm$ 0.007 & 0.01 & 0.1 \\
         MSNBC & -6.042 $\pm$ 0.003 & 0.01 & 0.1 & \textbf{-6.041} $\pm$ 0.001  & 0.01 & 0.1 \\
         KDDCup2k & -2.350 $\pm$ 0.001 & 0.001 & 0.1 & -2.360 $\pm$ 0.003  & 0.0001 & 0.1 \\
         Plants & -12.878 $\pm$ 0.026 & 0.01 & 0.1 & -12.908 $\pm$ 0.020  & 0.01 & 0.1 \\
         Audio & -39.841 $\pm$ 0.018 & 0.001 & $10^{-6}$ & -39.938 $\pm$ 0.038  & 0.001 & 0.1 \\
         Jester & -53.235 $\pm$ 0.031 & 0.001 & 0.1 & -53.234 $\pm$ 0.026 & 0.001 & $10^{-6}$ \\
         Netflix & -56.818 $\pm$ 0.027 & 0.001 & $10^{-6}$ & -56.844 $\pm$ 0.028  & 0.001 & 0.1 \\
         Accidents & \textbf{-28.895} $\pm$ 0.122 & 0.0001 & 0.1 & -29.114 $\pm$ 0.070  & 0.0001 & 0.01 \\
         Retail & -11.092 $\pm$ 0.047 & 0.0001 & 0.1 & -11.142 $\pm$ 0.017  & 0.0001 & 0.1 \\
         Pumsb-star & \textbf{-24.101} $\pm$ 0.088 & 0.0001 & 0.1 & -24.206 $\pm$ 0.085  & 0.0001 & 0.1 \\
         DNA & -83.674 $\pm$ 0.233 & 0.0001 & 0.1 & -83.798 $\pm$ 0.182  & 0.0001 & 0.1 \\
         Kosarek & -11.043 $\pm$ 0.035 & 0.0001 & 0.1 & -11.188 $\pm$ 0.037  & 0.0001 & 0.1 \\
         MSWeb & -9.847 $\pm$ 0.025 & 0.0001 & 0.1 & -10.015 $\pm$ 0.018 & 0.0001 & 0.1 \\
         Book & -34.334 $\pm$ 0.093 & 0.0001 & 0.1 & -34.428 $\pm$ 0.080 & 0.0001 & 0.1 \\
         EachMovie & -56.842 $\pm$ 0.078 & 0.0001 & 0.1 & -57.129 $\pm$ 0.069  & 0.0001 & 0.1 \\
         WebKB & -159.533 $\pm$ 0.227 & 0.0001 & 0.1 & -160.601 $\pm$ 0.331  & 0.0001 & 0.1 \\
         Reuters-52 & -87.932 $\pm$ 0.144 & 0.0001 & 0.1 & -88.400 $\pm$ 0.161  & 0.0001 & 0.1 \\
         20 Newsgrp. & -122.162 $\pm$ 0.138 & 0.0001 & 0.1 & -122.827 $\pm$ 0.121  & 0.0001 & 0.1 \\
         BBC & -248.293 $\pm$ 0.464 & 0.0001 & 0.1 & \textbf{-247.815} $\pm$ 0.248  & 0.0001 & 0.1 \\
         Ad & -18.539 $\pm$ 0.152 & 0.001 & 0.1 & -18.738 $\pm$ 0.403 & 0.01 & 0.01 \\
         \bottomrule
    \end{tabular}
    \caption{Performance of LearnSPN \cite{correia2020joints} on binary datasets.}
    \label{tab:lspn_bin_det}
\end{table}

\begin{table}[H]
    \centering
    \small
    \begin{tabular}{|l|c|c|c|c|c|c|}
    \toprule
        \multirow{2}{*}{\textbf{Dataset}} & \multicolumn{6}{c|}{\textbf{LearnSPN}} \\
        \cline{2-7}
         & \multicolumn{3}{c|}{\textbf{EM}} & \multicolumn{3}{c|}{\textbf{K-means}} \\
         \cline{2-7}
         & \textbf{Test LL} & \textbf{$p$} & \textbf{$\alpha$} & \textbf{Test LL} & \textbf{$p$} & \textbf{$\alpha$} \\
        \hline
         bank & -20.139 $\pm$ 0.030 & 0.01 & 0.1 & -20.277 $\pm$ 0.045 & 0.01 & 0.01 \\
         electricity & -11.229 $\pm$ 0.011 & 0.001 & 0.01 & -11.488 $\pm$ 0.010 & 0.0001 & 0.1 \\
         segment & -17.517 $\pm$ 0.081 & 0.01 & 0.1 & -17.663 $\pm$ 0.090 & 0.01 & 0.01 \\
         german & -22.720 $\pm$ 0.158 & 0.001 & 0.1 & -22.857 $\pm$ 0.174 & 0.001 & 0.1 \\
         vowel & -16.957 $\pm$ 0.079 & 0.01 & $10^{-6}$ & -17.086 $\pm$ 0.076  & 0.01 & $10^{-6}$ \\
         cmc & -9.850 $\pm$ 0.087 & 0.01 & 0.1 & -9.902 $\pm$ 0.128  & 0.01 & 0.1 \\
    \bottomrule
    \end{tabular}
    \caption{Performance of LearnSPN \cite{correia2020joints} on mixed datasets}
    \label{tab:lspn_mix_det}
\end{table}

\begin{table}[H]
    \centering
    \small
    \begin{tabular}{|l|c|c|c|c|c|c|}
        \toprule
        \multirow{2}{*}{\textbf{Dataset}} & \multicolumn{6}{c|}{\textbf{\softl}} \\
        \cline{2-7}
         & \multicolumn{3}{c|}{\textbf{EM}} & \multicolumn{3}{c|}{\textbf{K-means}} \\
         \cline{2-7}
         & \textbf{Test LL} & \textbf{$p$} & \textbf{$\alpha$} & \textbf{Test LL} & \textbf{$p$} & \textbf{$\alpha$} \\
        \hline
         NLTCS & -5.979 $\pm$ 0.006 & 0.01 & $10^{-6}$ & \textbf{-5.974} $\pm$ 0.002 & 0.01 & 0.01 \\
         MSNBC & -6.056 $\pm$ 0.003 & 0.01 & 0.01 & -6.048 $\pm$ 0.001 & 0.01 & 0.01 \\
         KDDCup2k & -2.372 $\pm$ 0.006 & 0.0001 & $10^{-6}$ & -2.345 $\pm$ 0.005 & 0.01 & $10^{-6}$ \\
         Plants & -12.643 $\pm$ 0.015 & 0.01 & 0.1 & \textbf{-12.572} $\pm$ 0.013 & 0.01 & $10^{-6}$ \\
         Audio & \textbf{-39.659} $\pm$ 0.023 & 0.01 & 0.1 & -40.346 $\pm$ 0.014 & 0.01 & 0.01 \\
         Jester & \textbf{-53.005} $\pm$ 0.041 & 0.01 & 0.1 & -53.545 $\pm$ 0.019 & 0.01 & 0.1 \\
         Netflix & \textbf{-56.491} $\pm$ 0.016 & 0.01 & 0.1 & -57.698 $\pm$ 0.015 & 0.01 & $10^{-6}$ \\
         Accidents & -30.092 $\pm$ 0.133 & 0.0001 & 0.1 & -29.544 $\pm$ 0.049 & 0.0001 & 0.01 \\
         Retail & -11.040 $\pm$ 0.021 & 0.0001 & 0.1 & \textbf{-10.887} $\pm$ 0.010 & 0.01 & $10^{-6}$ \\
         Pumsb-star & -24.818 $\pm$ 0.244 & 0.0001 & 0.1 & -28.397 $\pm$ 0.212 & 0.0001 & 0.01 \\
         DNA & -83.026 $\pm$ 0.240 & 0.0001 & 0.01 & \textbf{-82.062} $\pm$ 0.078 & 0.01 & $10^{-6}$ \\
         Kosarek & -11.108 $\pm$ 0.009 & 0.0001 & 0.1 & \textbf{-10.890} $\pm$ 0.020 & 0.01 & $10^{-6}$ \\
         MSWeb & -9.881 $\pm$ 0.049 & 0.001 & 0.1 & \textbf{-9.688} $\pm$ 0.007 & 0.001 & $10^{-6}$ \\
         Book & -34.173 $\pm$ 0.080 & 0.0001 & 0.1 & \textbf{-33.031} $\pm$ 0.022 & 0.01 & $10^{-6}$ \\
         EachMovie & -57.401 $\pm$ 0.166 & 0.0001 & 0.1 & -55.225 $\pm$ 0.058 & 0.0001 & $10^{-6}$ \\
         WebKB & -162.361 $\pm$ 1.400 & 0.0001 & 0.1 & -158.703 $\pm$ 1.015 & 0.01 & 0.01 \\
         Reuters-52 & -90.240 $\pm$ 0.525 & 0.001 & 0.1 & -88.338 $\pm$ 0.239 & 0.01 & 0.01 \\
         20 Newsgrp. & -121.218 $\pm$ 0.138 & 0.0001 & 0.1 & \textbf{-121.091} $\pm$ 0.127 & 0.01 & $10^{-6}$ \\
         BBC & -249.381 $\pm$ 0.937 & 0.0001 & 0.1 & -250.080 $\pm$ 0.533 & 0.01 & 0.01 \\
         Ad & -20.309 $\pm$ 0.794 & 0.01 & 0.01 & -40.129 $\pm$ 1.791 & 0.01 & 0.01 \\
         \bottomrule
    \end{tabular}
    \caption{Performance of \softl on binary datasets.}
    \label{tab:softl_bin_det}
\end{table}

\begin{table}[H]
    \centering
    \small
    \begin{tabular}{|l|c|c|c|c|c|c|}
    \toprule
        \multirow{2}{*}{\textbf{Dataset}} & \multicolumn{6}{c|}{\textbf{\softl}} \\
        \cline{2-7}
         & \multicolumn{3}{c|}{\textbf{EM}} & \multicolumn{3}{c|}{\textbf{K-means}} \\
         \cline{2-7}
         & \textbf{Test LL} & \textbf{$p$} & \textbf{$\alpha$} & \textbf{Test LL} & \textbf{$p$} & \textbf{$\alpha$} \\
        \hline
         bank & \textbf{-19.993} $\pm$ 0.044 & 0.0001 & 0.1 & -20.132 $\pm$ 0.033 & 0.01 & 0.01 \\
         electricity & \textbf{-11.217} $\pm$ 0.021 & 0.01 & $10^{-6}$ & -11.422 $\pm$ 0.012 & 0.0001 & 0.1 \\
         segment & \textbf{-17.480} $\pm$ 0.071 & 0.001 & $10^{-6}$ & -17.625 $\pm$ 0.100  & 0.001 & 0.01 \\
         german & -22.423 $\pm$ 0.186 & 0.01 & 0.1 & \textbf{-22.395} $\pm$ 0.187  & 0.01 & 0.1 \\
         vowel & \textbf{-16.584} $\pm$ 0.156 & 0.01 & $10^{-6}$ & -17.347 $\pm$ 0.169  & 0.01 & 0.01 \\
         cmc & -9.820 $\pm$ 0.098 & 0.01 & 0.01 & \textbf{-9.811} $\pm$ 0.060 & 0.001 & 0.01 \\
    \bottomrule
    \end{tabular}
    \caption{Performance of \softl on mixed datasets}
    \label{tab:softl_mix_det}
\end{table}

\section{Comparison with State of the Art}

\begin{table}[H]
    \centering
    \small
    \begin{tabular}{|l|c|c|c|c|c|c|c|c|}
        \toprule
        \multirow{2}{*}{\textbf{Dataset}} & \multicolumn{2}{c|}{\textbf{LearnSPN}} & \multirow{2}{*}{\textbf{CNET}} & {\textbf{Learn}} & \multirow{2}{*}{\textbf{HCLT}} & \multirow{2}{*}{\textbf{EiNet}} & \textbf{RAT} & {\textbf{Soft}} \\
        \cline{2-3}
         & \textbf{Gens} & \textbf{Correia} &  & \textbf{PSDD} &  &  & \textbf{SPN} & \textbf{Learn}\\
        \hline
         NLTCS & -6.110 & -5.995 & -6.10 & -6.03 & -5.99 & -6.015 & -6.01 & \textbf{-5.974} \\
         MSNBC & -6.113 & \textbf{-6.041} & -6.06 & -6.04 &  -6.05 & -6.119 & -6.04 & -6.048 \\
         KDDCup2k & \textbf{-2.182} & -2.350 & -2.21 & -2.12 & -2.18 & -2.183 & -2.13 & -2.345 \\
         Plants & -12.977 & -12.878 & -13.37 & -13.79 & -14.26 & -13.676 & -13.44 & \textbf{-12.572} \\
         Audio & -40.503 & -39.841 & -46.84 & -41.98 & -39.77 & -39.879 & -39.96 & \textbf{-39.659} \\
         Jester & -75.989 & -53.234 & -64.50 & -53.47 & \textbf{-52.46} & -52.563  &  -52.97 & -53.005 \\
         Netflix & -57.328 & -56.818 & -69.74 & -58.41 & \textbf{-56.27} &  -56.544 & -56.85 & -56.491 \\
         Accidents & -30.038 & -28.895 & -31.59 & -33.64 & \textbf{-26.74} & -35.594 & -35.49 & -29.544 \\
         Retail & -11.043 & -11.092 & -11.12 & \textbf{-10.81} & -10.84 & -10.916 & -10.91 & -10.887 \\
         Pumsb-star & -24.781 & \textbf{-24.101} & -25.06 & -33.67 & -23.64 & -31.954 & -32.53 & -24.818 \\
         DNA & -82.523 & -83.674 & -109.79 & -92.67 & \textbf{-79.05} & -96.086 & -97.23 & -82.062 \\
         Kosarek & -10.989 & -11.043 & -11.53 & -10.81 & \textbf{-10.66} & -11.029 & -10.89 & -10.890 \\
         MSWeb & -10.252 & -9.847 & -10.20 & -9.97 & -9.98 & -10.026 & -10.12 & \textbf{-9.688} \\
         Book & -35.886 & -34.334 & -40.19 & -34.97 & -33.83 & -34.739 & -34.68 & \textbf{-33.031} \\
         EachMovie & -52.485 & -56.842 & -60.22 & -58.01 & \textbf{-50.81} & -51.705 & -53.63 & -55.225 \\
         WebKB & -158.204 & -159.533 & -171.95 & -161.09 & \textbf{-152.77} & -157.282 & -157.53 & -158.703 \\
         Reuters-52 & \textbf{-85.067} & -87.932 & -91.35 & -89.61 & -86.26 & -87.368 & -87.37 & -88.338 \\
         20 Newsgrp. & -155.925 & -122.162 & -176.56 & -161.09 & -153.40 & -153.938 & -152.06 & \textbf{-121.091} \\
         BBC & -250.687 & \textbf{-247.815} & -300.33 & -253.19 & -251.04 & -248.332 & -252.14 & -249.381  \\
         Ad & -19.733 & -18.539 & -16.31 & -31.78 & 
         \textbf{-16.07} & -26.273 & -48.47 & -20.309  \\
         \bottomrule
    \end{tabular}
    \caption{Performance results of \softl vs. LearnSPN \cite{gens2013learning}, LearnSPN \cite{correia2020joints}, CNET \cite{rahman2014cutset}, LearnPSDD \cite{liang2017learning}, Hidden Chow Liu Trees \cite{NEURIPS2021_1d0832c4}, Einsum Networks \cite{pmlr-v119-peharz20a}, and RAT-SPN \cite{peharz2020random} over binary datasets.}
    \label{tab:res_bin_apndx}
\end{table}

Once again, we would like to note that our motivation for proposing \softl is not to compete with the state of the art methods, but to introduce a better \textit{base model} compared to LearnSPN. LearnSPN is utilized as a base model (or a building block) for many subsequent algorithms that managed to achieve impressive competitive results, outperforming LearnSPN itself. Such algorithms can utilize \softl interchangeably, and based on the results reported in this paper, we believe that utilizing \softl instead of LearnSPN will lead to noticeable performance gains, helping other algorithms achieve or surpass state of the art. Hence, our experiments were mainly designed to compare \softl with its main competitor, LearnSPN, on a variety of test configurations. Nevertheless, we include a comparison between \softl and some of the state of the art methods (namely LearnPSDD \cite{liang2017learning}, Hidden Chow Liu Trees \cite{NEURIPS2021_1d0832c4}, Einsum Networks \cite{pmlr-v119-peharz20a}, and RAT-SPN \cite{peharz2020random}) on binary datasets in table \ref{tab:res_bin_apndx}.

\softl manages to outperform LearnPSDD on 16 out of 20 datasets, EiNet on 14 out of 20 datasets, and RAT-SPN on 15 out of 20 datasets. The only method that performs better on average than \softl is HCLT (\softl still outperforms HCLT on 8 out of 20 datasets), which has a larger set of hyperparameters and is fine-tuned over a larger (and more precise) set of hyperparameter configurations. In addition, \softl manages to achieve the best results (among all the aforementioned methods) over 6 out of 20 datasets (On a separate not, we would like to mention that ID-SPN \cite{rooshenas2014} also manages to statistically outperform \softl on average, however, we do not consider ID-SPN a reasonable competitor since its resulting structures are orders of magnitude larger than the structures learned by \softl and LearnSPN. This means that ID-SPN potentially achieves better performance at the cost of over-parametrization and larger amount of computations, which does not lead to an unbiased comparison). These results show that despite its simple design as a base model (like LearnSPN), \softl is competitive to state of the art models, while having a very large room to grow when used as a building block of more elaborate methods.

\section{Theoretical Intuition}\label{a4}

In this section, we will provide theoretical insight into how methods like LearnSPN and \softl optimize the global likelihood. Before going into the details, we first reiterate the learning process of LearnSPN/\softl and establish necessary assumptions. In LearnSPN/\softl, each learning iteration does one of the following tasks: i) adding a (factorized) leaf node at the end of a path; ii) adding an internal sum node; iii) adding an internal product node.
We assume that we have the option to terminate LearnSPN/\softl at any arbitrary iteration of learning. Doing so means that we stop clustering over instances/variables and finalize the PC by adding factorized distributions to each path that does not already end in a leaf node. We call the resulting PC produced by this process the \textit{alternative PC} at iteration $t$. If we terminate the algorithm at the root of the PC, then the alternative PC would be a fully factorized distribution as in Figure \ref{fig:alt1}. Similarly, if we first add a sum node (with two children) to the root of the PC, and then proceed to terminate the learning, the alternative PC would have a structure similar to Figure \ref{fig:alt2}

\begin{figure}
    \centering
    \begin{subfigure}{0.35\textwidth}
        \centering
        \scalebox{0.6}{\definecolor{blu}{RGB}{199, 206, 234}
\definecolor{gre}{RGB}{181, 234, 215}
\definecolor{re}{RGB}{255, 154, 162}
\definecolor{ore}{RGB}{255, 218, 193}
\definecolor{lgr}{RGB}{226, 240, 203}
\definecolor{mel}{RGB}{255, 183, 178}

\begin{tikzpicture}[
    cross/.style={minimum size=8mm, fill=lgr,
        path picture={
            \draw[black] (path picture bounding box.south east) -- (path picture bounding box.north west) (path picture bounding box.south west) -- (path picture bounding box.north east);
        }
    },
    sum/.style={minimum size=8mm, fill=blu,
        path picture={
            \draw[black] (path picture bounding box.south) -- (path picture bounding box.north) (path picture bounding box.west) -- (path picture bounding box.east);
        }
    },
    integral/.style={minimum size=6mm, fill=re, 
        inner sep=0pt,
        text width=4mm,
        align=center,},
    gauss/.style={minimum size=8mm, fill=ore,
        path picture={
            \draw[black] plot[domain=-.2:.2] ({\x},{exp(-100*\x*\x -2.)});
        }
    },
    indicator/.style={minimum size=6mm, fill=ore,
        path picture={
            \draw[black] plot[domain=-.1:.1] ({\x},{((\x > 0) - 0.5)*0.35});
        }
    },
    dt/.style={minimum size=5mm, fill=white
    },
    line/.style={
      draw, thick,
      -latex',
      shorten >=2pt
    },
    cloud/.style={
      draw=red,
      thick,
      ellipse,
      fill=red!20,
      minimum height=1em
    },
    spn/.style={
    regular polygon,
    regular polygon sides=3,
    draw,
    fill=white
    }
]

    \node[draw, circle, cross] (p) at (0, 0) {};
    \node[draw, circle, gauss] (g1) at (-2, -2) {};
    \node[draw, circle, gauss] (g2) at (2, -2) {};
    \node[draw, circle, gauss] (g3) at (-1, -2) {};

    \node (ind) at (-2.25, -2.75) {$p^{0}_{1}(X_{1})$};
    \node (ind) at (-1.0, -2.75) {$p^{0}_{2}(X_{2})$};
    \node (ind) at (2.25, -2.75) {$p^{0}_{m}(X_{m})$};
    \node (ind) at (0, 0.75) {$p^{0}(X_{1}, ..., X_{m})$};
    \node (ind) at (0.5, -2) {$...$};

    \draw[line width=0.4mm, ->] (p) to (g1) {};
    \draw[line width=0.4mm, ->] (p) to (g2) {};
    \draw[line width=0.4mm, ->] (p) to (g3) {};

\end{tikzpicture}}
        \caption{Caption}
        \label{fig:alt1}
    \end{subfigure}
    \begin{subfigure}{0.55\textwidth}
        \centering
        \scalebox{0.6}{\definecolor{blu}{RGB}{199, 206, 234}
\definecolor{gre}{RGB}{181, 234, 215}
\definecolor{re}{RGB}{255, 154, 162}
\definecolor{ore}{RGB}{255, 218, 193}
\definecolor{lgr}{RGB}{226, 240, 203}
\definecolor{mel}{RGB}{255, 183, 178}

\begin{tikzpicture}[
    cross/.style={minimum size=8mm, fill=lgr,
        path picture={
            \draw[black] (path picture bounding box.south east) -- (path picture bounding box.north west) (path picture bounding box.south west) -- (path picture bounding box.north east);
        }
    },
    sum/.style={minimum size=8mm, fill=blu,
        path picture={
            \draw[black] (path picture bounding box.south) -- (path picture bounding box.north) (path picture bounding box.west) -- (path picture bounding box.east);
        }
    },
    integral/.style={minimum size=6mm, fill=re, 
        inner sep=0pt,
        text width=4mm,
        align=center,},
    gauss/.style={minimum size=8mm, fill=ore,
        path picture={
            \draw[black] plot[domain=-.2:.2] ({\x},{exp(-100*\x*\x -2.)});
        }
    },
    indicator/.style={minimum size=6mm, fill=ore,
        path picture={
            \draw[black] plot[domain=-.1:.1] ({\x},{((\x > 0) - 0.5)*0.35});
        }
    },
    dt/.style={minimum size=5mm, fill=white
    },
    line/.style={
      draw, thick,
      -latex',
      shorten >=2pt
    },
    cloud/.style={
      draw=red,
      thick,
      ellipse,
      fill=red!20,
      minimum height=1em
    },
    spn/.style={
    regular polygon,
    regular polygon sides=3,
    draw,
    fill=white
    }
]

    \node[draw, circle, sum] (r) at (0, 2) {};

    \node[draw, circle, cross] (p1) at (-3, 0) {};
    \node[draw, circle, gauss] (g11) at (-5, -2) {};
    \node[draw, circle, gauss] (g12) at (-1, -2) {};
    \node[draw, circle, gauss] (g13) at (-4, -2) {};

    \node (ind) at (-5.25, -2.75) {$p^{1}_{11}(X_{1})$};
    \node (ind) at (-4, -2.75) {$p^{1}_{12}(X_{2})$};
    \node (ind) at (-0.75, -2.75) {$p^{1}_{1m}(X_{m})$};
    \node (ind) at (-3.25, 0.75) {$p^{1}_{1}(X_{1}, ..., X_{m})$};
    \node (ind) at (-2.5, -2) {$...$};

    \node (ind) at (-1.5, 1.25) {$w_{1}$};
    \node (ind) at (1.5, 1.25) {$w_{2}$};
    \node (ind) at (0, 2.75) {$p^{1}(x_{1}, ..., X_{m})$};

    \draw[line width=0.4mm, ->] (p1) to (g11) {};
    \draw[line width=0.4mm, ->] (p1) to (g12) {};
    \draw[line width=0.4mm, ->] (p1) to (g13) {};

    \node[draw, circle, cross] (p2) at (3, 0) {};
    \node[draw, circle, gauss] (g21) at (1, -2) {};
    \node[draw, circle, gauss] (g22) at (5, -2) {};
    \node[draw, circle, gauss] (g23) at (2, -2) {};

    \node (ind) at (0.75, -2.75) {$p^{1}_{21}(X_{1})$};
    \node (ind) at (2, -2.75) {$p^{1}_{22}(X_{2})$};
    \node (ind) at (5.25, -2.75) {$p^{1}_{2m}(X_{m})$};
    \node (ind) at (3.25, 0.75) {$p^{1}_{2}(X_{1}, ..., X_{m})$};
    \node (ind) at (3.5, -2) {$...$};

    \draw[line width=0.4mm, ->] (p2) to (g21) {};
    \draw[line width=0.4mm, ->] (p2) to (g22) {};
    \draw[line width=0.4mm, ->] (p2) to (g23) {};
    \draw[line width=0.4mm, ->] (r) to (p1) {};
    \draw[line width=0.4mm, ->] (r) to (p2) {};

\end{tikzpicture}}
        \caption{Caption}
        \label{fig:alt2}
    \end{subfigure}
    \caption{Caption}
    \label{fig:alt}
\end{figure}

Depending on the task performed by the learning algorithm, the difference between any two consecutive alternative PCs (alternative PC at iteration $t$ and alternative PC at iteration $t+1$) can be: i) nothing (adding leaf nodes is inconsequential in alternative PCs since the process of termination includes adding a leaf node to any path without leaves); ii) a sum node; iii) a product node. If there is no difference between consecutive alternative PCs, then the likelihood of the data stays the same between the two PCs. In the case where the difference between two alternative PCs is in a product node, it can be easily deducted that the likelihood still remains the same, since a product node by itself (with factorized leaves) cannot induce any changes to the likelihood (the product of two factorized distributions over disjoint variables is equal to the product of univariate distributions over each variable). This leaves us only with the case where the difference between two consecutive alternative PCs is in a sum node. In this case (we can take the PCs represented in Figures \ref{fig:alt1} and \ref{fig:alt2} as an example; the same logic can be generalized to any arbitrary structure at any arbitrary iteration, with the exception that the difference in the likelihood will be weighted by a positive multiplier), the difference in the likelihood stems from the difference between $p^{0}(X_{1}, ..., X_{m})$ and $p^{1}(X_{1}, ..., X_{m})$ (or in the general case, the difference between $p^{t}(X_{1}, ..., X_{m})$ and $p^{t+1}(X_{1}, ..., X_{m})$), where $p^{0}(X_{1}, ..., X_{m})$ is a factorized distribution learned on some data $\mathcal{D}$, and $p^{1}(X_{1}, ..., X_{m})$ is a mixture of $C$ (number of output clusters) factorized distributions based on the resulting clusters, learned from the same data $\mathcal{D}$. If we can guarantee that $p^{1}$ has a better likelihood compared to $p^{0}$, then we can conclude that in every iteration of learnSPN/\softl, the likelihood either increases or stays the same, a process that gradually maximizes the likelihood of the data as the learning algorithm proceeds.

Whether or not $p^{1}$ is an increase over $p^{0}$ in terms of likelihood depends on the clustering algorithm. Yet, we can prove that there is always a solution that obtain the same likelihood as (or better than) the alternative option. Without losing generality, assume that we are adding a sum node with two children. In the case of soft clustering, one can equally divide each sample between the two clusters (weighing half to each side), which will lead to the same likelihood as the alternative fully factorized model. Hence, any soft clustering algorithm that performs a job at least as good as this (which is surely expected) obtains the same or better likelihood after the sum node is added. When using hard clustering, the argument is slightly different (as we cannot ``split'' the data points in half to use in each side). In this case, one can think of the following split: choose an arbitrary data point and take all its $t\geq 1$ copies to be represented in one of the child of the sum node (with weight $w$ equal to $t$ over the number of data points in consideration) and all other points to the other child (with weight $1-w$). It is not hard to see that both children will have better likelihood than the one of the alternative model for the part of the data points they represent, and thus the appropriate weighting yields a better model than the alternative model.

In addition to the intuitive perspective on clustering methods, we would also like to mention that some algorithms such as hard/soft EM, if properly initialized, can theoretically guarantee that the resulting mixture will have a better likelihood compared to the alternative factorized model.

\end{document}